\documentclass[sn-mathphys]{sn-jnl}% Math and Physical Sciences Reference Style
\jyear{2021}%

\theoremstyle{thmstyleone}%
\theoremstyle{thmstyletwo}%

\theoremstyle{thmstylethree}%

\usepackage[caption=false,font=footnotesize]{subfig}

\begin{document}

\title[Article Title]{Influence of collaborative customer service by service robots and clerks in bakery stores}


\author*[1,2]{\fnm{Yuki} \sur{Okafuji}}\email{okafuji\_yuki\_xd@cyberagent.co.jp}
\author[1,2]{\fnm{Sichao} \sur{Song}}
\author[1,2]{\fnm{Jun} \sur{Baba}}
\author[2]{\fnm{Yuichiro} \sur{Yoshikawa}}
\author[2]{\fnm{Hiroshi} \sur{Ishiguro}}

\affil*[1]{\orgdiv{AI Lab}, \orgname{CyberAgent, Inc.}, \orgaddress{\city{Shibuya} \postcode{150--6121}, \state{Tokyo}, \country{Japan}}}

\affil[2]{\orgdiv{Graduation School of Engineering Science}, \orgname{Osaka University}, \orgaddress{\city{Toyonaka} \postcode{560--0043}, \state{Osaka}, \country{Japan}}}


\abstract{
In recent years, various service robots have been introduced in stores as recommendation systems. Previous studies attempted to increase the influence of these robots by improving their social acceptance and trust. However, when such service robots recommend a product to customers in real environments, the effect on the customers is influenced not only by the robot itself, but also by the social influence of the surrounding people such as store clerks. Therefore, leveraging the social influence of the clerks may increase the influence of the robots on the customers. Hence, we compared the influence of robots with and without collaborative customer service between the robots and clerks in two bakery stores. The experimental results showed that collaborative customer service increased the purchase rate of the recommended bread and improved the impression regarding the robot and store experience of the customers. Because the results also showed that the workload required for the clerks to collaborate with the robot was not high, this study suggests that all stores with service robots may show high effectiveness in introducing collaborative customer service.
}




\maketitle

\section{Introduction}\label{sec1}
In recent years, service robots have been used widely in various scenarios; one such use of a service robot is as a recommendation system for sales promotion. These robots have been experimentally introduced in real commercial facilities and are expected to be a new recommendation system. For example, the service robots recommend products at shopping malls~\cite{Song21}, convenience stores~\cite{Kamei10}, department stores~\cite{Watanabe15}, and bakeries~\cite{Song22}, and experiments have demonstrated an increase in the corresponding product sales. The implementation of recommendation techniques using service robots has been currently verified in a limited number of situations; however, they are expected to be used widely if their utility is proven in a large number of cases.

When service robots recommend products, two main elements are important to strengthen the influence of the robots: social acceptance and trustworthiness of the robot~\cite{Ghazali20, Herse18a}. First, in order to recommend products to customers, the robot needs to initiate a conversation with them. However, service robots in a real environment are often ignored by customers even when they directly approach the customers and talk to them~\cite{Lee12, Tanaka16}. An interactive recommendation cannot be achieved if a conversation between the robot and customer is not initiated; consequently, the advantage of using a dialogue system, such as an interactive robot, disappears. Therefore, we must improve the social acceptance of the robots to achieve interactive recommendations. Next, to find information and products that customers actually prefer, and assist the customers in making purchase decisions, it is essential to build trust between the customers and all recommendation systems including but not limited to robots~\cite{Pu12}. In particular, when using a robot recommendation system, the previous study has reported that a high level of trust in a robot facilitates acceptance of recommendations from the robot~\cite{Rau09}. Improving the level of trust in the robot's suggestions helps in establishing a more effective recommendation system. 

Hence, previous studies have focused on investigating various methods to improve the social acceptance and trustworthiness of the robot itself as a recommendation system. Meanwhile, in this study, we assume that these service robots will not be introduced in an environment where only robots serve the customers but in stores that are already operated by human staff. In this case, the social acceptance and trustworthiness of the robot may be influenced not only by the robot itself but also by other people around the robot, including the working staff. Such influences from the surroundings are called social influences, and are considered not only in human--robot interaction (HRI)~\cite{Malhotra99} but also in human--human interaction (HHI)~\cite{Cialdini04}. Although the topic of service operations involving robots and human staff has been discussed~\cite{Belanche20, Xiao19}, it is still unclear what benefits can be gained by the influence of the staff through field experiments.

Accordingly, we assumed that the social acceptance and trustworthiness of robots, which strengthen their influence on the customers, are affected by the surrounding people in a real environment. Therefore, the aim of this study is to improve the social acceptance and trustworthiness of robots as recommendation systems by leveraging social influences and to improve the influence on customers through the improved social acceptance and trustworthiness. In particular, we create an atmosphere in which robots have already been accepted by collaborating customer service (CCS) with store clerks. This is a simple method, but CCS may be able to improve the social acceptance and trustworthiness of the robot from the perspective of the customer. Consequently, the improved influence of the robot owing to high social acceptance and trust can promote sales. In addition, we consider that collaboration with clerks can be adopted universally and it improves the persuasive recommendation by robots proposed in previous studies. Therefore, in this study, we introduced two service robots in two bakery stores. A field experiment was conducted to verify how the recommendation effect of the robot was influenced by CCS between the robot and store clerks.

The remainder of this paper is organized as follows. The related works are described in Section~\ref{sec2}. The experimental methodology and results are presented in Sections~\ref{sec3} and~\ref{sec4}, respectively. In Section~\ref{sec5}, a discussion based on the results and the scope for future studies are presented. Finally, Section~\ref{sec6} presents our conclusions.

\section{Related Works}\label{sec2}
In order to utilize robots as recommendation systems in the real world, the social acceptance and trustworthiness of robots are important~\cite{Ghazali20, Herse18a}. This is because robots are often ignored by customers in the real world~\cite{Lee12, Tanaka16} and improving the trustworthiness of robots improves persuasiveness~\cite{Rau09}. On the other hand, many studies aiming at developing robot recommendation systems directly verify the persuasiveness of robots, which means that the users' decisions can be influenced by robots, rather than the social acceptance or trustworthiness of robots. Therefore, we mainly present the literature on the persuasiveness of robots which is especially related to the trustworthiness.

Many studies have focused on nonverbal behavior to improve the persuasiveness of robots. In particular, it has been found that a robot's gaze and gesture expressions improve its persuasiveness~\cite{Ham15}. These nonverbal cues influence its persuasiveness more than simple voice-based communication~\cite{Chidambaram12}. Although simple voice-based communication has limited influence on the persuasiveness, the effects of various types of persuasive verbal communications have been verified. For example, because expertise improves the persuasiveness in HHI~\cite{Cialdini01}, robots can utilize expert knowledge and rhetorical communication~\cite{Andrist13}. In addition, the persuasiveness of a robot is enhanced through emotional verbalization~\cite{Saunderson22, Bertacchini17} and by utilizing the user's estimated emotions and social media information~\cite{Bertacchini17}.

The abovementioned studies have focused on improving the persuasiveness of a single robot, but a few studies have attempted the same with multiple robots. In general, people tend to agree with the opinions of others in their surroundings, which is called normative conformity. This concept was experimentally verified in HHI~\cite{Asch56}. Similarly, people also tend to agree with the opinions of multiple robots~\cite{Salomons17}. The persuasiveness of robots can be also improved by the balance theory~\cite{Heider58}. In a group with two robots and one human, it has been shown that adjusting the relationship between the two robots to balance the group state increases the persuasiveness of the robot toward the human~\cite{Kadowaki08}. In addition, another study using multiple robots showed that sequential persuasion, in which multiple robots talk repeatedly to customers, enhances persuasiveness, and the level of persuasiveness increases with an increase in the number of robots~\cite{Tae21}. Thus, methods to improve the persuasiveness of robots have been verified from various perspectives.

Some studies have utilized the persuasiveness of these robots in recommendation systems. In general, in the context of recommendation, the trustworthiness has been shown to shape source credibility and improve persuasion~\cite{Okeefe90, Haiman49}. Therefore, in robots as well, several studies have focused on the trustworthiness and persuasiveness of robots as recommendation systems. The basic persuasiveness of robots has been investigated in terms of their embodiment~\cite{Herse18a} and proactive recommendation ability~\cite{Peng19}. Many recommendation systems have been proposed to estimate the user's state during the interaction and recommend products according to the estimated state~\cite {Herse18a, Carolis17, Woiceshyn17, Alslaity19}. For example, the user's state such as attitude, emotion, and product preferences was estimated from the voice, posture, and facial expressions, and products were recommended based on these assessments in a dress shopping scenario~\cite{Carolis17}. The long-term adaptability of personalized recommendation systems by estimating user preferences through multiple interactions has been improved~\cite{Woiceshyn17}. Thus, the aim of previous studies was to develop a more persuasive recommendation system by personalizing the system using various methods (e.g.,~\cite{Alslaity19}).

The studies introduced in the previous paragraph verified the effectiveness of the robot recommendation system through laboratory experiments. Whereas some studies have demonstrated the practicality of recommendation systems as a new sales promotion method by verifying their effectiveness in real environments. As an example, the successful sale of toothbrushes by a teleoperated robot was demonstrated in a commercial facility~\cite{Song21}, and a recommendation robot in a bakery improved the store sales~\cite{Song22}. Three autonomous robots introduced in a convenience store promoted the sale of the recommended products~\cite{Kamei10}. It has also been shown that the longitudinal recommendation effect increases by adding confidence and word--of--mouth to the recommendation statement~\cite{Okafuji21}. Further, an android robot realized approximately twice as much product sales as the average human salesperson~\cite{Watanabe15}. In addition, robot systems that provide users with coupons and tastings that indirectly influence sales promotions have also been proposed~\cite{Shiomi13, Okafuji22, Tonkin17}. These studies examined the robot's size~\cite{Shiomi13}, types of robot behavior~\cite{Okafuji22}, and aggressiveness of the robot's talk~\cite{Tonkin17}. Studies on robot recommendation systems have also compared the performance of humans and virtual agents~\cite{Shiomi13, Okafuji22, Tonkin17, Herse18b, Sakai22}, and some studies have shown that the performance of the robot system exceeds human performance~\cite{Okafuji22}. However, in these studies, the persuasiveness of robots which is especially related to the trustworthiness is mainly evaluated rather than the social acceptance of robots, even though they are conducted in a real environment.

Studies have been conducted on the various aspects of using robots as recommendation systems. However, although they have directly examined the persuasiveness of robots to influence users' decisions, few studies examined in detail the social acceptance and trustworthiness that influence their persuasiveness. In addition, although a few studies have focused on the persuasiveness of multiple robots~\cite{Kadowaki08, Salomons17, Tae21}, most have examined the effect of recommendation by a single robot.  As described in the introduction, we assumed that these service robots will be introduced in stores that are already operated by humans. In this case, the social influence of the surrounding people, including store clerks, may also affect the relationship between the robot and customers. In the hospitality field, the topic of three-way interaction among customers, service robots, and employees has been recently discussed~\cite{Belanche20, Xiao19}. However, the actual effect of such interactions has not yet been examined. Therefore, it is necessary to investigate in detail the influence of CCS when introducing service robots at a real retailer.

\section{Methodology}\label{sec3}
\subsection{Overview}
The aim of this study is to investigate whether the social acceptance and trustworthiness of robots in the service field are improved and whether the recommendation effect is further improved through CCS between robots and clerks. To verify this, we installed service robots that recommended products at two bakery stores and conducted field experiments. The experiment was conducted for 15 days from 14--28 April 2022, and the robot was operated between 10 am -- 2 pm and 3 pm -- 5 pm. We informed all customers regarding the experiment and the related videotaping through a notification board. The study was conducted on an opt-out basis for unwilling customers who did not wish to be videotaped. 

This study was approved by the Research Ethics Committee of Osaka University (Reference number: R1-5-9).

\subsection{Robot Systems}
We developed a teleoperated robot system that enabled an operator to remotely control the robot and talk through it via a web-based video calling application. The system comprised three main components: a robot controller application, an operator interface, and a server. The social robot used in this system was ``Sota,'' developed by Vstone Co., Ltd. Sota is approximately 0.3 m tall, and it has functions such as voice and LED-generated facial expressions. Operators can monitor the video sent by the robot and speak accordingly through the robot to the customers in real time. The operator can control the gaze direction and some gestures of the robot using the interface.

\subsection{Experimental fields}
\begin{figure}[!t]
    \begin{center}
    \includegraphics[width=10.0cm]{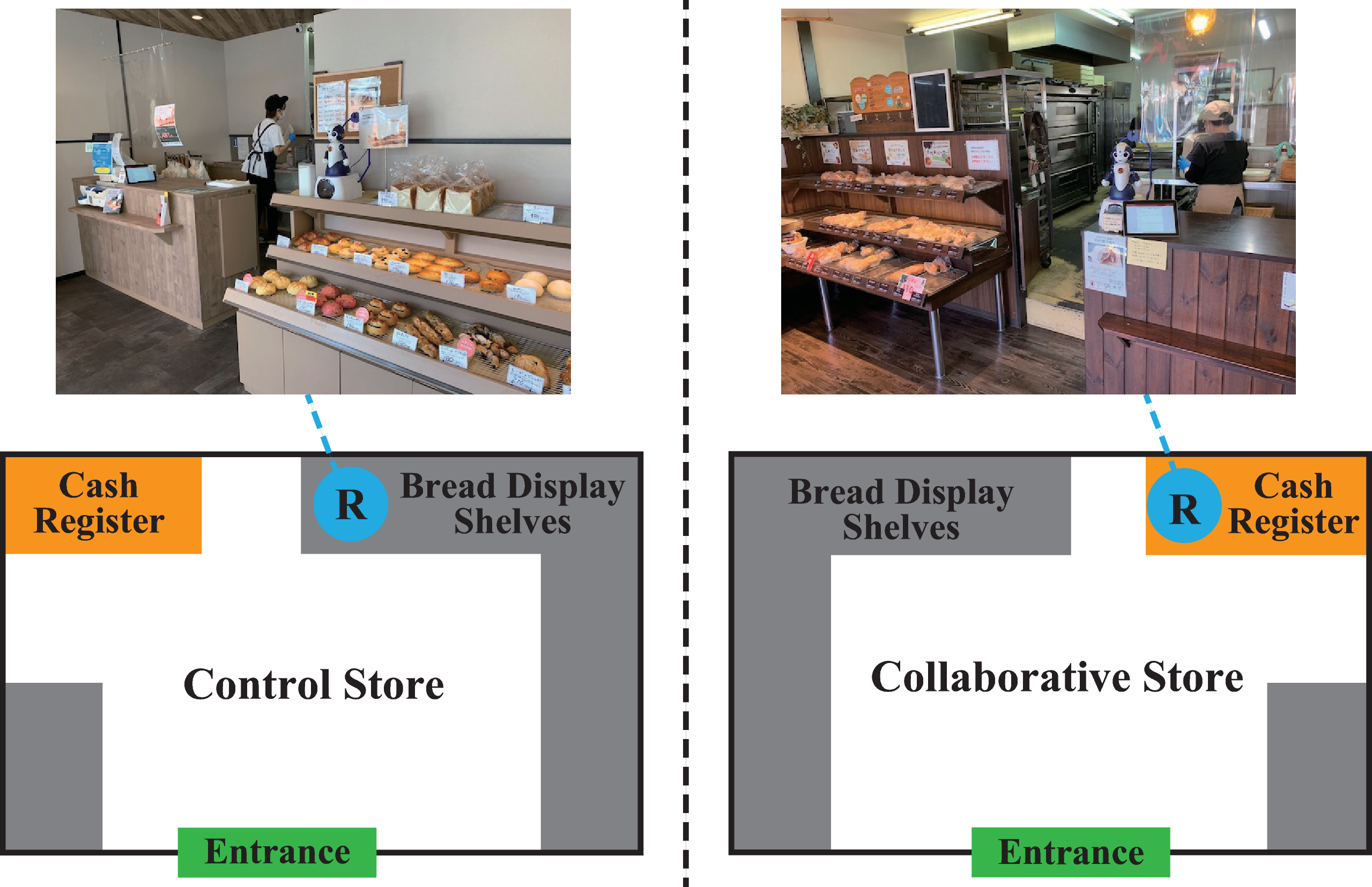}
    \caption{Layouts of two bakery stores and experimental setups. The left and right figures show the layouts of the Control and Collaborative Stores, respectively.}
    \label{fig:bakery}
    \end{center}
\end{figure}

We conducted experiments simultaneously in two bakeries with similar environments, selected from multiple candidates, to evaluate CCS. These bakery stores are located in Osaka and Kyoto, Japan, in residential areas outside the city center. Apart from small organizations such as elementary schools, there are no large universities, hospitals or businesses nearby. They sell approximately 50 types of breads. Fig.~\ref{fig:bakery} shows the layout of each store and the experimental setup.

\subsection{Robot Tasks}

One robot was installed in each store. The main tasks of the robot were to recommend specific bread products, converse with customers, and announce events such as the availability of freshly baked bread. Welcoming customers and promoting products are listed as tasks that retailers expect robots to perform~\cite{Niemela17a}; these tasks were included in this experiment. In terms of recommendations, the robot provided information about specific products in detail to invoke the interest of the customer. Regarding the products for recommendation, we discussed with the shop owners and selected five from approximately 50 kinds of breads at each shop.

As verbal behavior, the robot was able to provide and answer questions about detailed information about specific bread since using expertise improves persuasiveness~\cite{Andrist13}. In addition, in order to make recommendations according to the user's state~\cite{Carolis17}, the robot recommended different specific bread in accordance with the position and behavior of the customer. As nonverbal behavior, the robot always gazes at the customer and expresses gestures~\cite{Ham15}. The gesture expressions include pointing to bread, raising hands, waving the body, and so on. These gesture expressions are automatically generated when approximately 50 types of registered words such as greetings are spoken by operators.

Four robot operators, all females in their twenties, were employed for the tasks. They are all voice actors and skilled performers. The operators spoke in a voice by modulating their own voices similar to that of a small child to maintain consistency with the childlike appearance of the robot. This teleoperated approach has been referred to as ``Wizard of Oz (WoZ)''~\cite{Riek12}. Two of the four operators were assigned to each store daily on a random basis to eliminate differences in the operator performance.

The WoZ style was applied in this experiment because if an autonomous robot is used, communication errors might occur, and it will be difficult to evaluate the effectiveness of CCS. In addition, we instructed the operators to not admit to customers that they were human even if they were asked. If customers know that the robot is operated by humans, they would evaluate customer service for the human operator instead of the robot. To perform the experiment correctly, we informed the store clerks at each store that the robot was operated by humans. However, we instructed them not to divulge this information to customers. Therefore, this experimental environment created a scenario wherein the clerks were aware that they worked with a robot-mediated human whereas the customers were given the impression that they were being served by an autonomous robot. These experimental settings have been used in another study~\cite{Song22}.

\subsection{Clerk Operations}
For evaluating CCS, the robot tasks in the two bakeries were the same but the clerks' operations were different. As a control condition, CCS was not applied in one bakery, and the robot attempted to serve the customers alone. The clerks communicated information to the robot only when freshly baked bread was available; otherwise, they provided customer service as usual without involving the robot.

At the other bakery, the robot and clerks provided CCS. Specifically, we generated situations in which the customer was always presented with interactions between the robot and clerks while the customer was visiting the store. For example, the clerk repeated the statement made by the robot: 
\begin{itemize}
    \item Robot: \textit{``I recommend the salt bread today!''}
    \item Clerk: \textit{``We recommend the salt bread today!''}
\end{itemize}
The clerk responds to a statement from the robot:
\begin{itemize}
    \item Robot: \textit{``Customers have come!''}
    \item Clerk: \textit{``Okay, Thanks! Welcome!''}
\end{itemize}
The clerk chats with the robot:
\begin{itemize}
    \item Robot: \textit{``What is your recommendation?''}
    \item Clerk: \textit{``It's the salt bread.''}
    \item Robot: \textit{``Hey customers! The salt bread seems to be recommended!''}
\end{itemize}
We suppose that these interactions convey an atmosphere to the customers that the robot is accepted by the store clerks.

In addition, such collaboration is considered to strengthen two types of communication: sociality and support giving/getting~\cite{Fay11a}. Informal communication strengthens employee satisfaction~\cite{Fay11b} and relationships~\cite{Alparslan15}; thus, clerks may easily accept robot-mediated operators. Improving the relationship between the robot and clerks will have a more positive influence on the customer's impression regarding the robot.

Hereafter, the store without CCS is referred to as the Control Store, and the store with CCS is referred to as the Collaborative Store, as shown in Fig.~\ref{fig:bakery}.

\subsection{Measurement}
In this study, we evaluated the following: purchase rates of the recommended breads and questionnaires presented to customers, clerks, and operators.

First, the sales of the five types of recommended bread were evaluated to verify the influence of the robot. It was difficult to compare the sales between the two stores because the types of bread sold in each store and the recommended breads were different. Therefore, within each store, the influence of the robot was verified by comparing the purchase rates of the recommended bread when the robot was in operation and when it was not in operation. The opening hours of the Control and Collaborative Stores were 7 am to 6 pm and 7 am to 7 pm, respectively. Therefore, the robots operated for six hours (10 am -- 2 pm and 3 pm -- 5 pm) in each store and recommended specific types of bread, whereas they did not operate for five hours at the Control Store and six hours at the Collaborative Store each day. When not in operation, the robot remained inside the store, but all movements were stopped. Therefore, the influence of the robot was evaluated by comparing the following equation during the period of its operation and non-operation within each store:
\begin{equation}
    Purchase\ Rate\ (PR)\ [\%] = \frac{N_{Rec}}{N_{All}} \times 100,
    \label{eq:PR}
\end{equation}
where $N_{Rec}$ indicates the units sold for five types of recommended breads and $N_{All}$ denotes the units sold for all types of breads.

Next, we surveyed the customers and clerks regarding their impressions of the robot and the store experience. The customer questionnaire was presented at the cash register, and all customers through the cash register were prompted to answer the questionnaire. All questionnaires were evaluated using a 1--7 grade Likert scale.

The customer questionnaire was related to the impression regarding the robot because the customers did not know that the robot was controlled by human operators. Seven items were listed in the questionnaire: intelligent, useful, easy to talk, enjoyable, friendly, influential, and trustworthy. These are part of the social acceptance model for robots~\cite{Heerink10}. In addition, installing a robot may improve the impression regarding the store experience. For example, it has been reported that installing robots in hotels improves their brand experience~\cite{Chan19}. Therefore, we also evaluated the impression regarding the store experience under the influence of CCS. The following four items were listed in relation to the store experience: 1) ``Were you satisfied with this shop?'' 2) ``Did the robot improve the brand image of the store?'' 3) ``Did you feel anxious about robots usurping your job?'' and 4) ``Do you hope to revisit this store?''

Regarding the clerk questionnaire, we asked the participating clerks to answer the questionnaire every day. The clerk questionnaire asked about their impressions regarding the robot similar to the customer questionnaire. However, the clerk questionnaire may convey the impression regarding the human operator, not the robot, because the clerks were aware that the robot was operated by humans. The items listed in the impression regarding the robot were the same as those in the customer questionnaire. The four items listed in the impression regarding the store experience were 1) ``Did the robot improve the impression of the store?'' 2) ``Did you feel anxious about robots usurping your job?'' 3) ``Did you feel frustrated about working with the robot?'' and 4) ``Do you hope to work with the robot again?''

In HHI, humans build rapport through communication, including informal communication\cite{Alparslan15}. However, it is known that humans' perception of speaking objects is distorted when the appearance of the object is changed, even if they know that the object is operated by humans~\cite{Bennett17}. In addition, teleoperated humanoid robots are recognized as having lower intelligence than humans even when they perform the same task~\cite{Kuwamura12}. Therefore, it is unclear whether the relationships between clerks and robot-mediated operators are established in the same way as in HHI. Hence, the relationship between these variables was investigated using the clerk questionnaire.

Finally, we administered a questionnaire to operators regarding their workload, motivation, and sense of belonging. It is known that informal communication makes people develop a sense of belonging, even during telework, which increases their commitment to the work~\cite{Fay11a}. Therefore, we evaluated whether similar results could be obtained in this experimental environment, even though the operator questionnaire was not directly related to the purpose of this study. The items in the operator questionnaire were based on NASA-TLX~\cite{Hart88}, and included the following: ``Did you have high motivation throughout today's work'' and ``Did you feel like a member of the store during today's work?''

\section{Results}\label{sec4}
We analyzed the PRs of the five types of recommended breads, the impressions regarding the robot and store experience from the customers and clerks, and work performance of the operators when the robot was placed in the two bakeries for 15 days. However, the experiment was canceled on 21st April at the Control Store because of poor physical condition of one of the operators. Thus, the Control Store was evaluated for 14 days. The customer questionnaires at the Control and Collaborative Stores received 216 and 98 responses, respectively (average: 15.4 and 6.5, respectively, per day). The numbers of responses to the clerk questionnaire at the Control and Collaborative Stores were 32 and 43, respectively (average: 2.3 and 2.9, respectively, per day).

Demographic information of customers in the two stores is listed in Tables~\ref{tab:number} and \ref{tab:group}. During the experiment period, we randomly selected 2 weekdays and 2 weekends at each store, and we annotated user information from 10 am to 5 pm. The demographic information between the two stores (305 and 543 customers in the Control and Collaborative Stores, respectively) shows no large difference.

\begin{table}[!t]
  \begin{center}
  \caption{Demographic information on observed customer age and gender.}\label{tab:number}
  \begin{tabular}{@{}ccccc@{}}
  \toprule
  & Number of & Number of & Number of & Number of \\
  Stores & male customers & female customers & adult customers & child customers \\
  & (Ratio \%) & (Ratio \%) & (Ratio \%) & (Ratio \%) \\ \midrule
  Control & 102 (33.4) & 203 (66.6) & 282 (92.5) & 23 (7.5) \\
  Collaborative & 185 (34.1) & 358 (65.9) & 487 (89.7) & 56 (10.3) \\
  \botrule
  \end{tabular}
  \end{center}
\end{table}

\begin{table}[!t]
  \begin{center}
  \caption{Demographic information on customer group.}\label{tab:group}
  \begin{tabular}{@{}ccc@{}}
  \toprule
  Stores & Ratio of solo customer \% & Ratio of group customers \% \\ \midrule
  Control & 72.2 & 27.8 \\
  Collaborative & 72.0 & 28.0 \\
  \botrule
  \end{tabular}
  \end{center}
\end{table}

\subsection{Purchase Rate of Recommended Bread}
\begin{figure}[!t]
    \begin{center}
    \includegraphics[width=8.0cm]{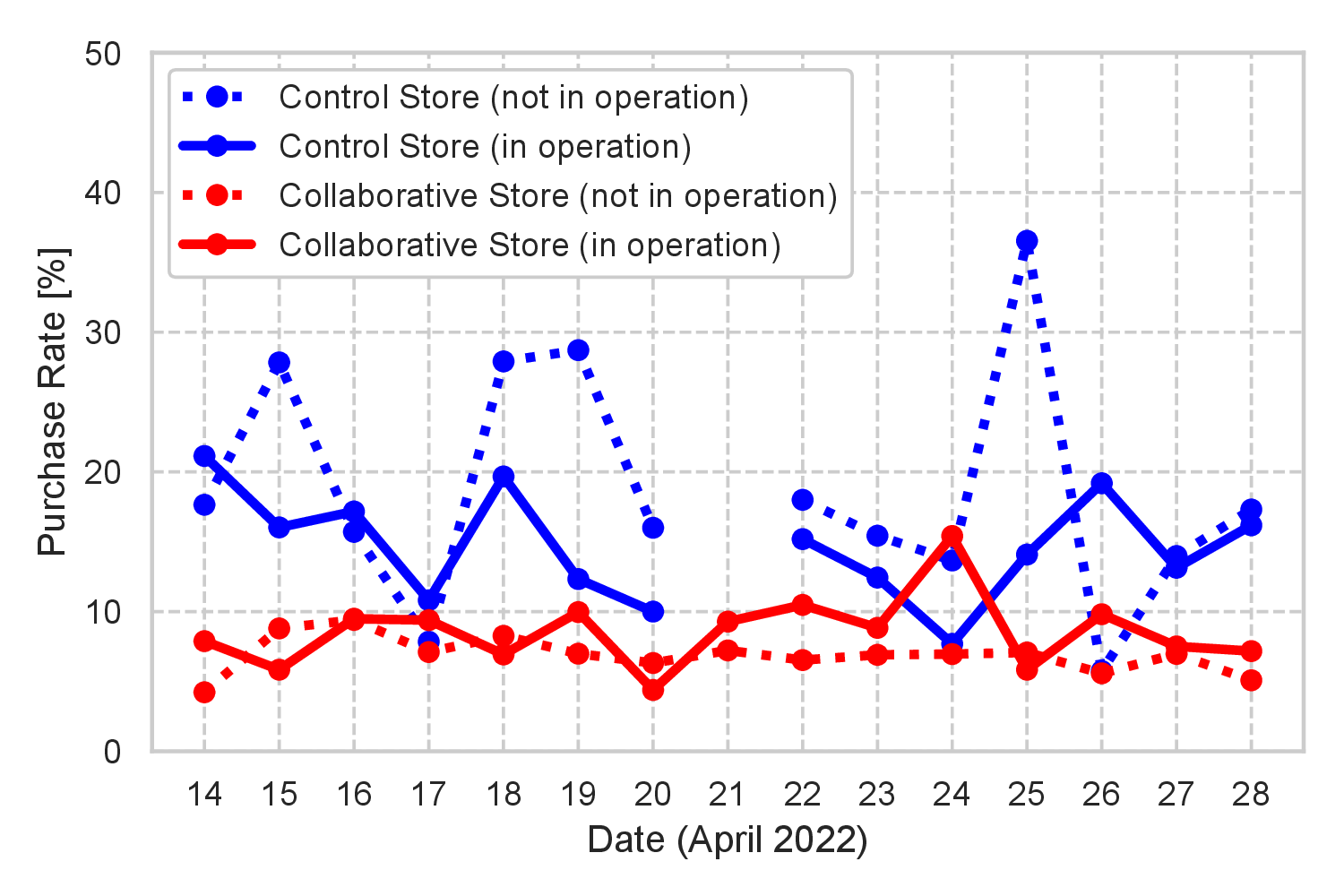}
    \caption{Purchase rate (PR) at each store.}
    \label{fig:sales}
    \end{center}
\end{figure}

Fig.~\ref{fig:sales} shows the PRs at both bakeries, as represented by Eq.~\ref{eq:PR}, divided into periods when the robot was in operation and when it was not in operation. In the Control Store, the average PR when the robot was not in operation was 15.5\% (264/1707), whereas the average PR when the robot was in operation was 14.5\% (345/2373). This means that the PR was reduced by approximately 6.0\% when the robot was in operation. However, in the Collaborative Store, the average PR when the robot was not in operation was 7.1\% (305/4305), whereas the average PR when the robot was in operation was 8.6\% (400/4637). This means that the PR increased by approximately 21.8\% when the robot was in operation. These results were verified within each store by using the chi-squared test. We used Cramer's $V$ as the effect size in the chi-squared test. The results for the Control Store revealed no significant differences in the PRs with and without the robot recommendation: ($\chi^2(1) = 0.67,~p = 0.71,~V = 0.01$). However, the results in the Collaborative Store revealed significant differences in the PR with and without the robot recommendation: ($\chi^2(1) = 7.30,~p = 0.02,~V = 0.03$).

\subsection{Customer Questionnaire}
\begin{figure}[!t]
    \begin{center}
    \includegraphics[width=8.0cm]{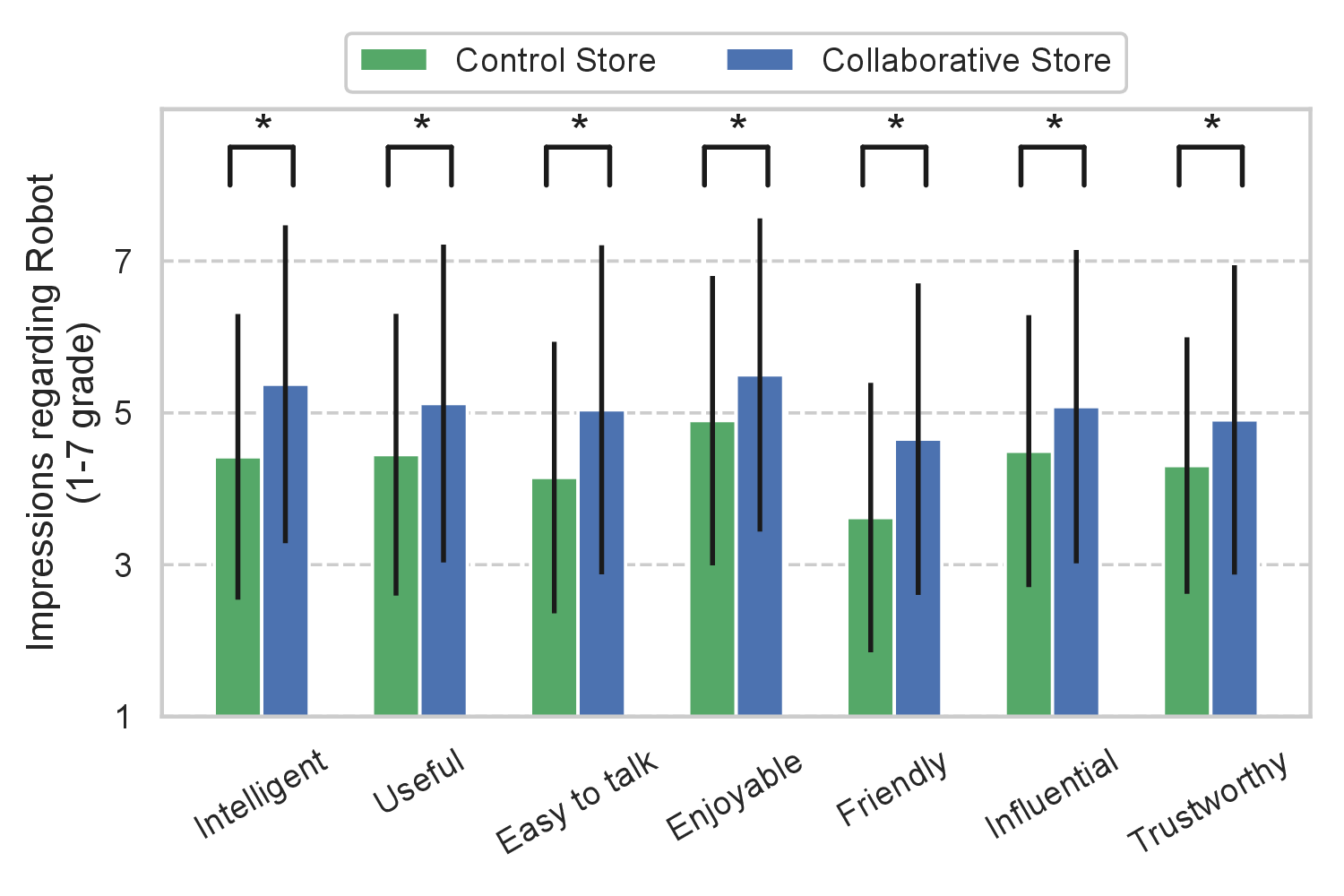}
    \caption{Impressions formed by customers regarding the robot. $*$ means $p < 0.05$.}
    \label{fig:CustomerRobot}
    \end{center}
\end{figure}

Fig.~\ref{fig:CustomerRobot} shows the results of the customers' impressions regarding the robot calculated by 216 and 98 customer questionnaires in the Control and Collaborative Stores, respectively. The value of 4 is neutral because of the use of the 1--7 grade Likert scale. Thus, the results show positive impressions regarding the robot in both stores, except for the item ``Friendly'' in the Control Store. In addition, the robot in the Collaborative Store gained more positive impressions from the customers in all aspects than the robot in the Control Store. These results were verified using the Mann--Whitney $U$ test. The results revealed significant differences in all items between the stores (\textit{Intelligent}: $U = 7530,~p < 0.01$, \textit{Useful}: $U = 8312,~p < 0.01$, \textit{Easy to talk}: $U = 7903.5,~p < 0.01$, \textit{Enjoyable}: $U = 8337,~p < 0.01$, \textit{Friendly}: $U = 7458,~p < 0.01$, \textit{Influential}: $U = 8385.5,~p < 0.01$, and \textit{Trustworthy}: $U = 8262.5,~p < 0.01$).

\begin{figure}[!t]
    \begin{center}
    \includegraphics[width=8.0cm]{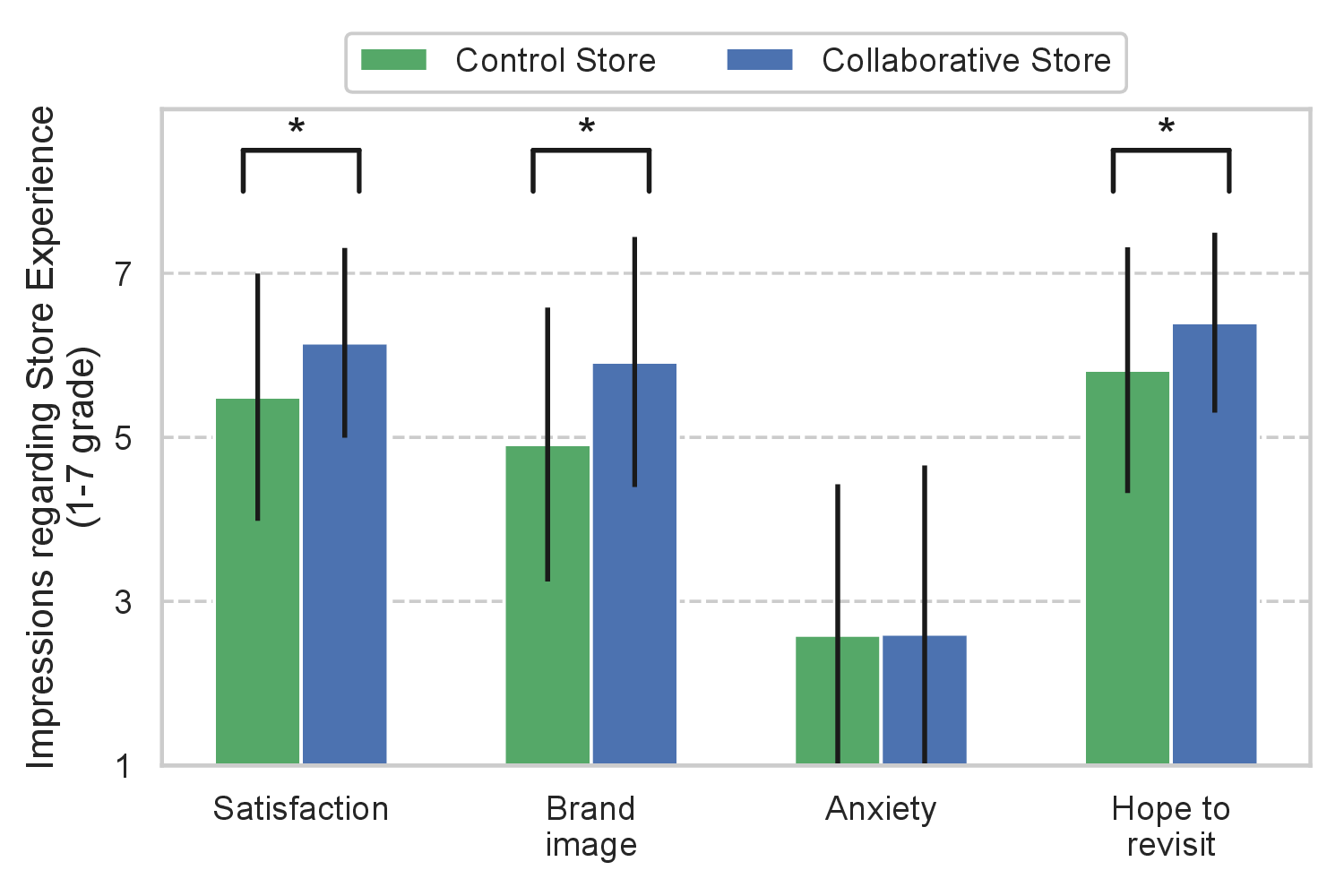}
    \caption{Impressions formed by customers regarding the store experience. $*$ means $p < 0.05$.}
    \label{fig:CustomerStore}
    \end{center}
\end{figure}

Fig.~\ref{fig:CustomerStore} shows the results of the customers' impressions regarding the store experience. The results show that the values for the Collaborative Store are greater for all items than those for the Control Store. The results revealed significant differences between the two stores (\textit{Satisfaction}: $U = 7894,~p < 0.01$, \textit{Brand image}: $U = 6786.5,~p < 0.01$, \textit{Hope to revisit}: $U = 8206.5,~p < 0.01$), with the exception of Anxiety ($U = 10140,~p= 0.53$).

\subsection{Clerk Questionnaire}
\begin{figure}[!t]
    \begin{center}
    \includegraphics[width=8.0cm]{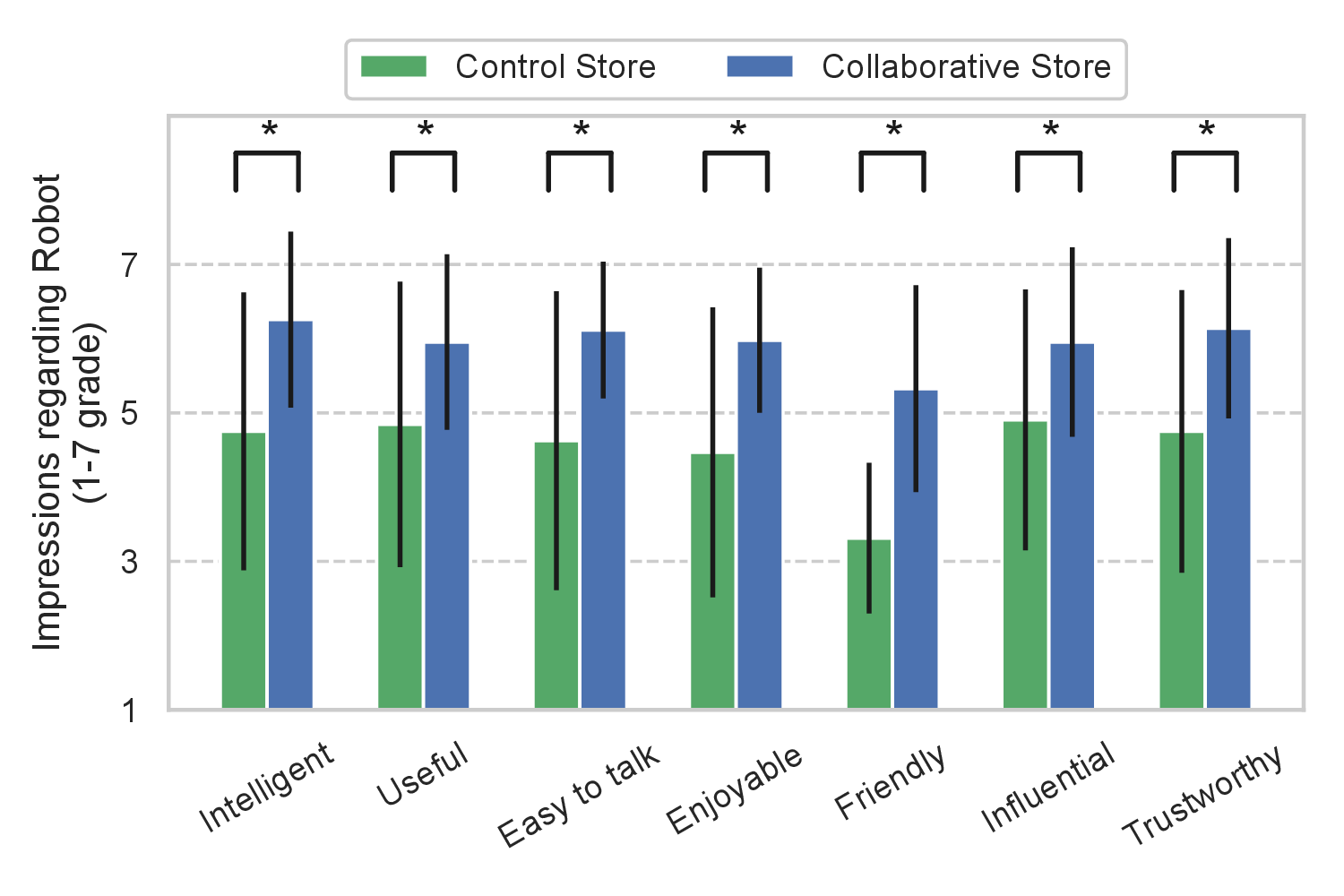}
    \caption{Impressions formed by clerks regarding the robot. $*$ means $p < 0.05$.}
    \label{fig:ClerkRobot}
    \end{center}
\end{figure}

Fig.~\ref{fig:ClerkRobot} shows the clerks' impressions regarding the robot calculated by 32 and 43 customer questionnaires in the Control and Collaborative Stores, respectively. The results show positive impressions of the robot in both stores, except for the item ``Friendly'' in the Control Store. In addition, the robot in the Collaborative Store received better impressions from the clerks in all aspects than the robot in the Control Store. These trends are the same as those of the results for the customers. These results were verified using the Mann--Whitney $U$ test, which revealed significant differences in all items between both stores (\textit{Intelligent}: $U = 407,~p < 0.01$, \textit{Useful}: $U = 489,~p = 0.03$, \textit{Easy to talk}: $U = 428,~p < 0.01$, \textit{Enjoyable}: $U = 395,~p < 0.01$, \textit{Friendly}: $U = 157.5,~p < 0.01$, \textit{Influential}: $U = 467,~p = 0.01$, and \textit{Trustworthy}: $U = 415.5,~p < 0.01$).

\begin{figure}[!t]
    \begin{center}
    \includegraphics[width=8.0cm]{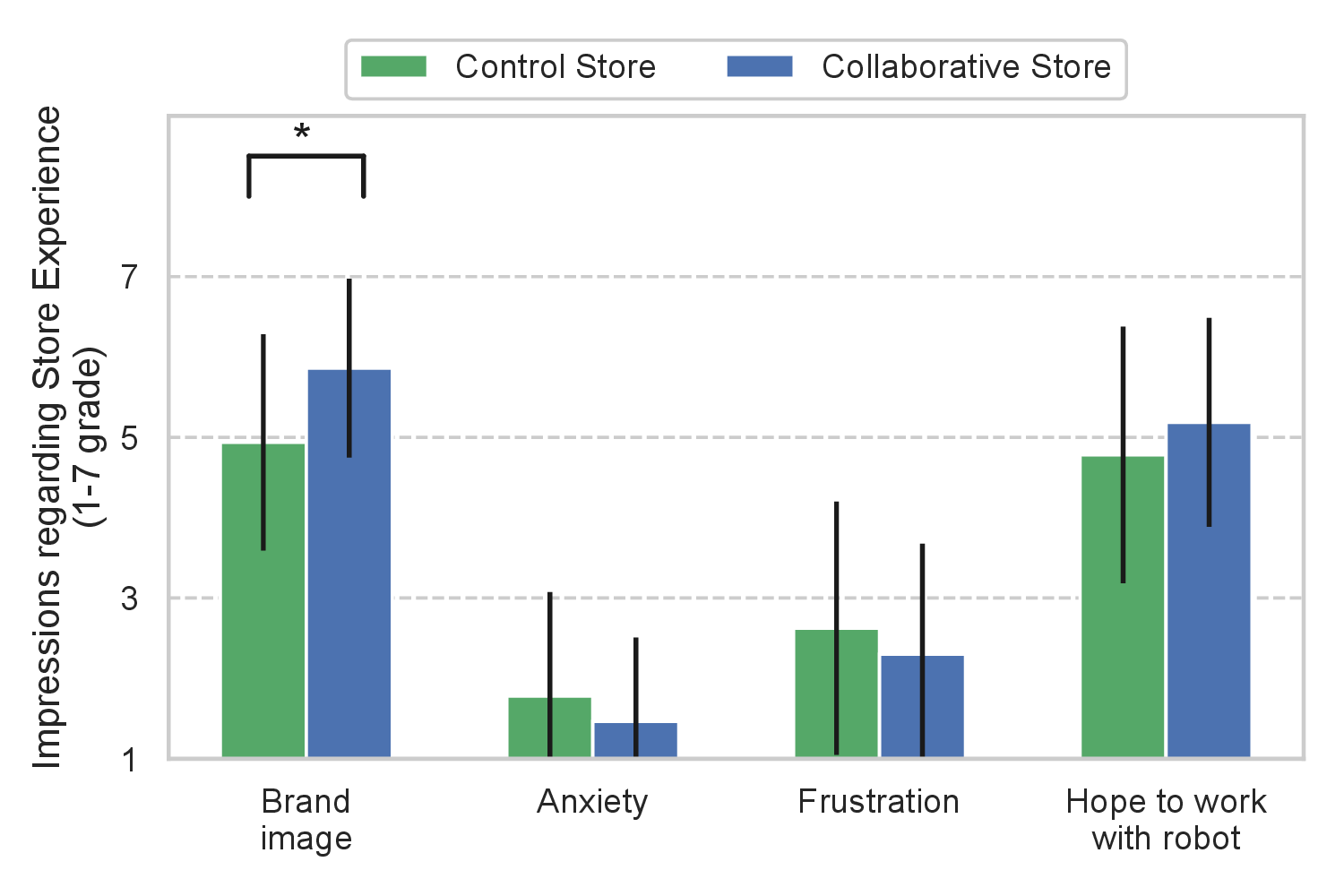}
    \caption{Impressions formed by clerks regarding the store experience. $*$ means $p < 0.05$.}
    \label{fig:ClerkStore}
    \end{center}
\end{figure}

Fig.~\ref{fig:ClerkStore} shows the results of the clerks' impressions regarding the store experience. The results for the Collaborative Store were greater for ``Brand image'' and ``Hope to work with robot'' than the corresponding values for the Control Store, and the opposite results were observed for ``Anxiety'' and ``Frustration.'' The results of both stores revealed significant differences in ``Brand image'' ($U = 426.5,~p < 0.01$) and no significant differences in other items (\textit{Anxiety}: $U = 617.5,~p = 0.30$, \textit{Frustration}: $U = 616,~p = 0.42$, and \textit{Hope to work with robot}: $U = 567.5,~p = 0.18$).

\subsection{Operator Questionnaire}
\begin{figure}[!t]
    \begin{center}
    \includegraphics[width=8.0cm]{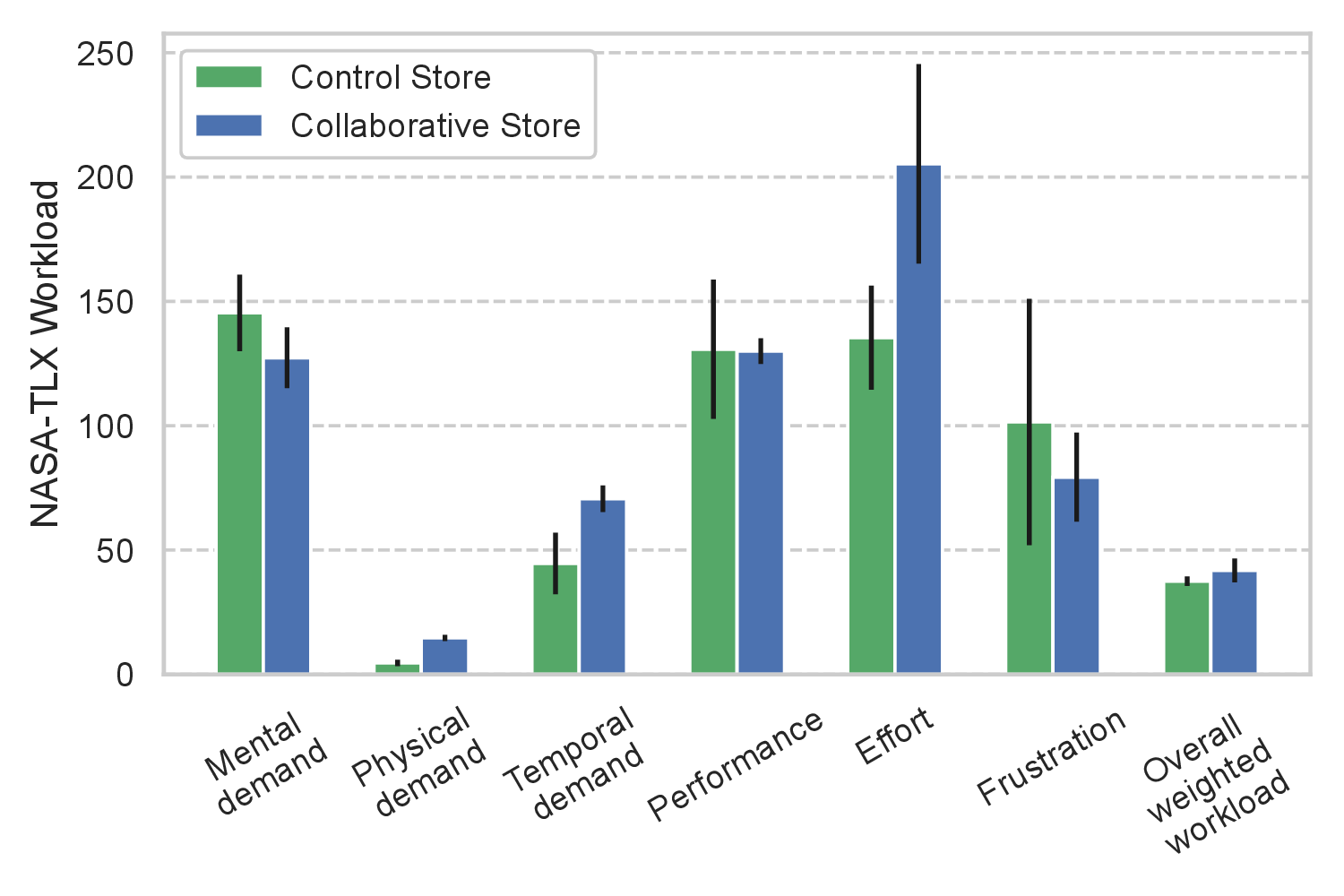}
    \caption{Workload results of the operators using NASA-TLX.}
    \label{fig:OperatorNASA}
    \end{center}
\end{figure}

Fig.~\ref{fig:OperatorNASA} shows the results of the questionnaire related to the operators' workload using NASA-TLX. The items ``Mental demand, '' ``Physical demand, '' ``Temporal demand, '' ``Performance, '' ``Effort, '' and ``Frustration'' are independent subscales, and ``Overall weighted workload' indicates the operator's overall workload calculated in subscales. These results were verified using the Mann--Whitney $U$ test, and the results revealed no significant differences in all items between both stores (\textit{Mental demand}: $U = 83,~p = 0.50$, \textit{Physical demand}: $U = 67,~p = 0.13$, \textit{Temporal demand}: $U = 76,~p = 0.32$, \textit{Performance}: $U = 95.5,~p = 0.93$, \textit{Effort}: $U = 64,~p = 0.12$, \textit{Frustration}: $U = 75,~p = 0.30$, and \textit{Overall weighted workload}: $U = 83,~p = 0.50$).

\begin{figure}[!t]
    \begin{center}
    \includegraphics[width=8.0cm]{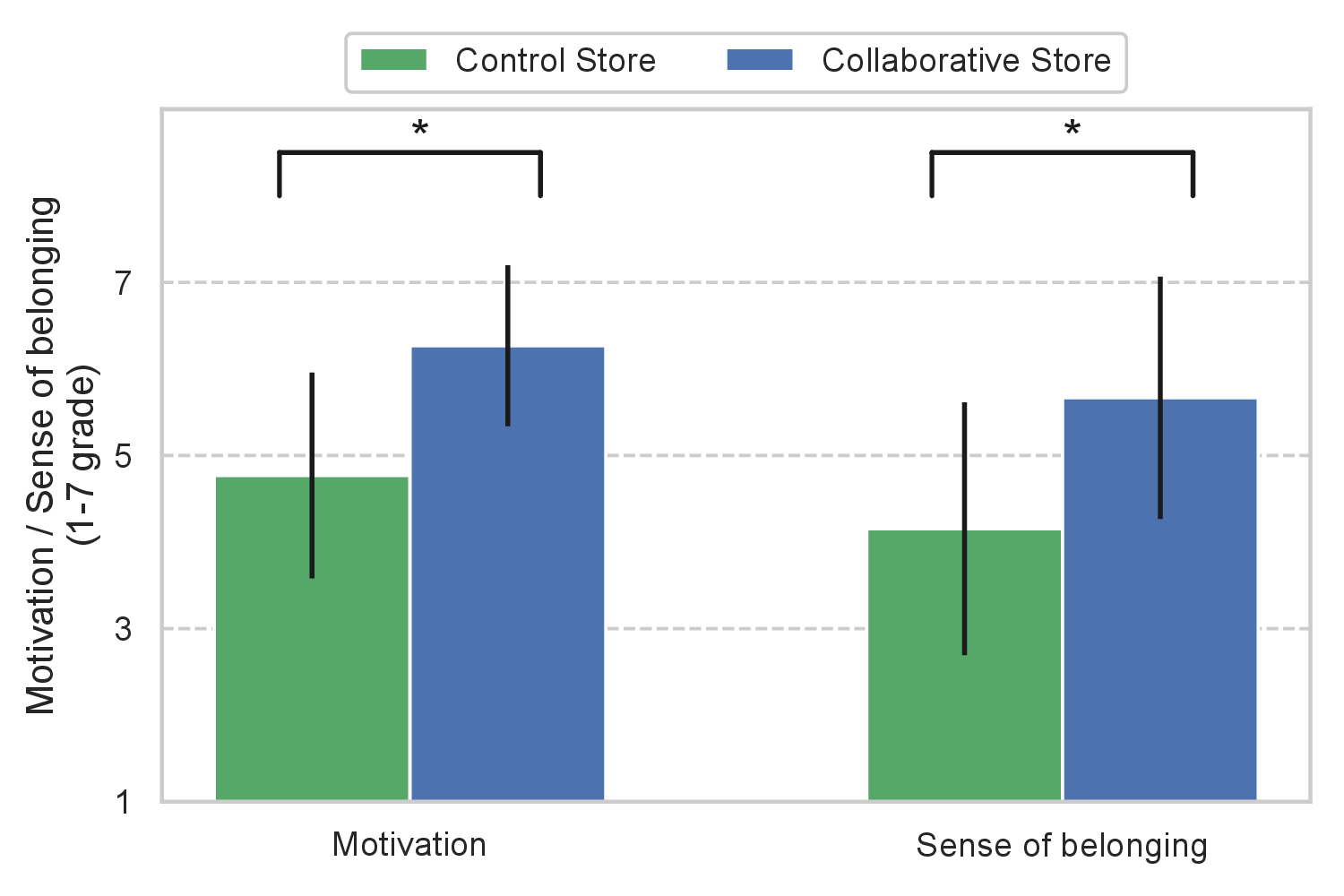}
    \caption{Motivation and sense of belonging of operators. $*$ means $p < 0.05$.}
    \label{fig:OperatorPerform}
    \end{center}
\end{figure}

Fig.~\ref{fig:OperatorPerform} shows the results of the operators' motivation and sense of belonging. Both results show that the values for the Collaborative Store were greater than those for the Control Store. The results revealed significant differences in both items between the stores (\textit{Motivation}: $U = 34,~p < 0.01$ and \textit{Sense of belonging}: $U = 42.5,~p < 0.01$).

\section{Discussion}\label{sec5}

\subsection{Effect of Collaborative Customer Service on Customers}
The results revealed no significant difference in the PR in the Control Store but showed a difference in the PR in the Collaborative Store when comparing the situations where the robot was in operation and not in operation. These results indicate that the use of the robot as a recommendation system had a greater influence on the customers in the Collaborative Store. The robots in both the stores were operated by the same operators and used the same strategy to recommend various types of bread. The Collaborative Store, where the clerks collaborated with the robot, aimed to create an atmosphere in which a relationship between the clerks and the robot was already established. In other words, the results suggest that the persuasiveness of the robot was influenced by the social influence of the clerks, and this influence affected the sales of the recommended types of bread.

Even though the manner of interaction of the robot with the customers was the same in both stores, customers' impressions of the robot improved for all items in the Collaborative Store. These items include the factors of social acceptance~\cite{Heerink10} and trust in the robot. Therefore, the effectiveness of the robot's recommendations was strengthened because the customers accepted and trusted the robot owing to social influence, which is consistent with the previous study~\cite{Rau09}. This is also similar to the phenomenon of social influence described in HHI~\cite{Cialdini04}. The results of this study imply that social influence of the clerks exists even in the use of service robots in the retail industry.

A comparison of the PR between the first and last halves of the experimental period showed the longitudinal influence of the robots. In the Control Store in the first eight days, the PR when the robot was in operation decreased by approximately 20.3\% when compared with the PR when the robot was not in operation. In contrast, the PR when the robot was in operation in the last seven days increased by 8.8\% when compared with the PR when the robot was not in operation. In the Collaborative Store, the PR increased by 7.5\% in the first half of the period, whereas it increased by 42.2\% in the last half of the period owing to the influence of the robot's recommendation.

Interestingly, in both stores, the influence of the robot was stronger during the last half of the period. In other words, it is possible that the greater the frequency of contact with the robot, the greater its influence, which is known as the mere exposure effect~\cite{Zajonc68}. In order to evaluate the customer behavior change, we annotated the number of customers who interacted with the robot during the first two days and the last two days of the experiment period for each store. In the Control Store, 16.3\% (14/86) and 19.0\% (16/84) of the customers interacted with the robot during the first and last two days, respectively. Whereas, in the Collaborative Store, 31.2\% (44/141) and 47.1\% (66/140) of the customers interacted with the robot during the first and last two days. In other words, the percentage of customers who interacted with the robot increased in the last half of the experiment in both stores. However, this is an indirect result to verify the long-term influence of the robot, and it is still unclear whether the impression regarding the robot was strictly improved in the longitudinal duration. This is because we could not collect the questionnaire from customers who interacted with the robot more than once. These results suggest that CCS between the clerks and the robot may have accelerated the establishment of the relationship between the robot and customers. If a long-term experiment can be conducted to fully establish a robot--customer relationship, the influence of the robot between the Control and Collaborative Stores could be comparable.

The results also show that CCS improves the customers' impressions of the store experience. In terms of brand image, this result is consistent with the results of a previous study~\cite{Chan19}. In contrast, one possible reason why the results for ``Anxiety'' did not show a significant difference between both stores is that the robot did not perform skillful tasks that made the user feel anxious. In this study, the robot's tasks were designed to include welcoming customers and promoting products, as proposed in a previous study~\cite{Niemela17a}, that is, they were not essential tasks (such as baking bread, displaying the bread, and handling the checkout) for the clerks working at the bakeries. We believe that because these support tasks do not give the impression of replacing human labor, they did not cause a difference in ``Anxiety'' between the two stores.

Further, the impressions regarding the robot and store experience were positive for many items, not only in the Collaborative Store but also in the Control Store. This implies that the introduction of service robots creates a positive impression on the customers, which is consistent with the results of previous studies~\cite{Niemela17b, SongCS22}. Therefore, the introduction of robots is beneficial in aspects other than sales, regardless of CCS.

\subsection{Influence of Collaborative Customer Service on Clerks and Operators}
The impression formed by clerks regarding the robot can be interpreted as an evaluation of the human operator rather than the robot. In this situation, the impression regarding the robot in the Collaborative Store showed better results for all items than that in the Control Store. The increased communication, including informal communication, between the robot and clerks, had a more positive effect on the robot. This is consistent with the results reported in previous studies on HHI~\cite{Fay11b, Alparslan15}. In addition, human perception of a speaking object is distorted even if it is known that the object is operated by another human~\cite{Bennett17, Kuwamura12}. However, we found that communicating with a robot-mediated human could produce the same positive impressions as in the normal HHI situation.

Regarding the workload of the operations, the Collaborative Store required collaboration between the clerks and the robot operators, which could have increased their workload. However, at both stores, the clerk questionnaire showed no significant difference for the item ``Frustration''; the operator questionnaire also showed no significant difference in the overall weighted workload of NASA-TLX. This implies that the application of CCS in the Collaborative Store did not have a significant effect on the workload, which suggests that CCS may not increase the workload of clerks but can increase the influence of the robot.

In addition, the evaluation of operators showed that CCS improved their motivation and sense of belonging. These results for robot-mediated telework are similar to those of a previous study on typical telework~\cite{Fay11a}.

\subsection{Problems of Collaboration between Humans and Autonomous Service Robots}
In this study, the experimental design used a robot operated by human operators, and clerks were aware of this setting. Therefore, it was easy to build a relationship between the clerks and robots, as we focused on evaluating CCS without the communication errors caused by autonomous robots. In addition, we expect that these service robots will be autonomous robots in the future, whereupon it is unclear whether they can build a similar relationship with the clerks~\cite{Lu19, Qiu20}. This is likely to become a major issue as service robots become more widespread in the future.

In studies on industry robots, human--robot collaboration has been actively examined~\cite{Ajoudani18}, as they have been prevalent before the introduction of service robots. One topic that is often discussed in human--robot collaboration is the relationship between robots and workers. Because industry robots have replaced human labor in factories, it has been pointed out that in some cases, employees do not welcome the introduction of robots~\cite{Welfare19}. Therefore, robots must be assigned to activities that do not diminish the preferences and satisfaction of humans. It has been shown that employees prefer to collaborate with robots in such environments~\cite{Welfare19, Ferreira14}. The same issue is expected to occur with service robots in the future~\cite{McClure18}.

Therefore, when introducing autonomous service robots in the future, we should pay attention to the relationship between robots and workers. In particular, we should eliminate the fear of workers regarding the introduction of service robots~\cite{McCartney20}. Involving workers in co-designing the human--robot collaboration can be effective in enabling workers to gain an understanding of service robots and not feel anxious regarding their jobs~\cite{Niemela19}. This study designed welcoming customers and promoting specific products as the tasks of the robot, but these are the expectations for robots that were answered by workers through workshops~\cite{Niemela17a}. Therefore, the actual introduction of service robots for performing these tasks may not be perceived positively by the workers. Thus, several considerations are required for tasks to be assigned to service robots~\cite{Niemela17c, Shi16}. We believe that the relationship between workers and service robots is likely to become an important issue in the future in a wide range of fields.

\subsection{Limitations}
This study had some limitations. First, we selected two bakery stores with similar environments from several candidate stores to verify the influence of CCS. However, these two stores did not have strictly identical environments such as the quality of customers. In addition, we compared the PRs during periods when the robot was in operation and not in operation. Because we had no pre-experiment sales data for recommended bread for each store by time of day we were unable to eliminate the effect of time of day on PRs from our analysis results. In other words, the tendency of the recommended bread purchased by the time of day may vary. Thus, we cannot argue that the results of this study can be strictly attributed to CCS. We also considered conducting A/B testing using a single store. In this case, it would be difficult to verify the effect of CCS because the results would be greatly influenced by the user's experience. In fact, this study showed that the influence of the robot improved in the last half of the experiment. Thus, we used two bakery stores to conduct this study to eliminate the factor of users' experience. This problem is not limited to this study but applies to many studies with wild field experiments. We need to discuss the extent to which rigorous comparisons are required in future field experiments.

Second, the long-term influence of the robot was not investigated in this study, as the longitudinal influence of the robot was examined for only 15 days. Previous studies on the social acceptance of robots spanned several months, for example, six months~\cite{Graaf16}. Therefore, 15 experimental days was not sufficient to investigate the long-term effect. The results of this study show that the influence of the robot improved in the last half of the experiment; however, it is unclear whether the social acceptance and trustworthiness of the robot also improved because we cannot collect the questionnaire from customers who interacted with the robot more than once. Thus, it is also unclear whether this effect will be maintained over several months. We expect that the long-term relationship between the robot and customers will be maintained if we continue using the WoZ robot. However, using an autonomous robot is likely to reduce the long-term influence of the robot.

Finally, the introduction of autonomous service robots was not considered in this study. As discussed in the previous section, the introduction of autonomous service robots into a store has many potential issues such as the tasks to be assigned to the robots and functions to be considered to build a relationship with the store clerks. Because all stakeholders have different expectations regarding the introduction of service robots~\cite{Niemela17a, McCartney20}, it is important to carefully discuss with these stakeholders and co-design autonomous robots before introducing them into the workplace~\cite{Niemela19}.

\section{Conclusion}\label{sec6}
The aim of this study was to increase the influence of robots as a recommendation system by leveraging the social influence of clerks. Therefore, we compared the influence of robots with and without collaborative customer service (CCS) with clerks in two bakery stores.

The results showed that CCS increased the purchase rate of the recommended types of bread and improved the impression regarding the robot and store experience of the customers. In addition, the impression formed by the clerks regarding the robot was also improved. Because the results showed that the workload for the collaboration between the robot and clerks was not high, this study suggests that all stores with service robots may show high effectiveness in introducing CCS.

In this study, WoZ robots were deployed in bakery stores to investigate the effects of CCS. However, these robots will operate autonomously in the future. In this case, various potential issues need to be considered, such as the tasks to be assigned to the robots and functions needed to build relationships with customers and workers.

\section*{Declarations}
\begin{description}
  \item[\bfseries{Authors' contributions}] YO, SS, and JB conceptualized the study. All the authors designed the experiments. SS and JB programmed the software. YO collected and analyzed the data. All authors interpreted the results. YO wrote the manuscript.
  \item[\bfseries{Funding}] This study was funded by CyberAgent, Inc.
  \item[\bfseries{Data Availability}] The datasets generated and/or analyzed during the current study are available from the corresponding author on reasonable request.
  \item[\bfseries{Conflict of interest}] This study received funding from the company CyberAgent, Inc. Authors YO, SS, and JB are employees of CyberAgent, Inc., and the remaining authors are employed at Osaka University. The funder was not involved in the study design, collection, analysis, interpretation of data, writing of this article, or the decision to submit it for publication. The authors declare that they have no other competing interests.
\end{description}






\begin{thebibliography}{1}
    \bibitem{Song21}
        Song, S, Baba, J, Nakanishi, J, Yoshikawa, Y, Ishiguro, H
        (2021)
        Teleoperated Robot Sells Toothbrush in a Shopping Mall: A Field Study.
        In: Extended Abstracts of the 2021 CHI Conference on Human Factors in Computing Systems
        221: 1--6,
        \url{https://doi.org/10.1145/3411763.3451754}
        
    \bibitem{Kamei10}
        Kamei, K, et al
        (2010)
        Recommendation from robots in a real-world retail shop.
        In: International Conference on Multimodal Interfaces and the Workshop on Machine Learning for Multimodal Interaction
        19: 1--8,
        \url{https://doi.org/10.1145/1891903.1891929}
        
    \bibitem{Watanabe15}
        Watanabe, M, Ogawa, K, Ishiguro, H
        (2015)
        Can Androids Be Salespeople in the Real World?.
        In: Annual ACM Conference Extended Abstracts on Human Factors in Computing Systems
        5: 251–-262,
        \url{https://doi.org/10.1007/s12369-013-0180-4}
        
    \bibitem{Song22}
        Song, S, Baba, J, Nakanishi, J, Yoshikawa, Y, Ishiguro, H
        (2022)
        Service Robots in a Bakery Shop: A Field Study.
        In: IEEE/RSJ International Conference on Intelligent Robots and Systems
        xxx--xxx,
        \url{https://doi.org/xxx} (to appear)
        
    \bibitem{Ghazali20}
        Ghazali, AS, Ham, J, Barakova, E, Markopoulos, P
        (2020)
        Persuasive Robots Acceptance Model (PRAM): Roles of Social Responses Within the Acceptance Model of Persuasive Robots.
        International Journal of Social Robotics
        12:1075--1092,
        \url{https://doi.org/10.1007/s12369-019-00611-1}
        
    \bibitem{Herse18a}
        Herse, S, et al
        (2018)
        Do You Trust Me, Blindly? Factors Influencing Trust Towards a Robot Recommender System.
        In: IEEE International Symposium on Robot and Human Interactive Communication
        7--14,
        \url{https://doi.org/10.1109/ROMAN.2018.8525581}
    
    \bibitem{Lee12}
        Lee, MK, Kiesler, S, Forlizzi, J, Rybski, P
        (2012)
        Ripple effects of an embedded social agent: a field study of a social robot in the workplace.
        In: SIGCHI Conference on Human Factors in Computing Systems
        User Modeling and User-Adapted Interaction
        695--704,
        \url{https://doi.org/10.1145/2207676.2207776}
        
    \bibitem{Tanaka16}
        Tanaka, K, Yamashita, N, Nakanishi, H, Ishiguro, H
        (2016)
        Teleoperated or autonomous?: How to produce a robot operator's pseudo presence in HRI.
        In: ACM/IEEE International Conference on Human Robot Interaction
        133--140,
        \url{https://doi.org/10.1109/HRI.2016.7451744}

    \bibitem{Pu12}
        Pu, P, Chen, L, Hu, R
        (2012)
        Evaluating recommender systems from the user's perspective: survey of the state of the art.
        User Modeling and User-Adapted Interaction
        22:317--355,
        \url{https://doi.org/10.1007/s11257-011-9115-7}
      
    \bibitem{Rau09}
        Rau, PLP, Li, Y, Li, D
        (2009)
        Effects of communication style and culture on ability to accept recommendations from robots.
        Computers in Human Behavior
        25(2):587--595,
        \url{https://doi.org/10.1016/j.chb.2008.12.025}
        
    \bibitem{Malhotra99}
        Malhotra, Y, Galletta, DF
        (1999)
        Extending the technology acceptance model to account for social influence: theoretical bases and empirical validation.
        In: Annual Hawaii International Conference on Systems Sciences
        1--14,
        \url{https://doi.org/10.1109/HICSS.1999.772658}
    
    \bibitem{Cialdini04}
        Cialdini, RB, Goldstein, NJ
        (2004)
        Social Influence: Compliance and Conformity.
        Annual Review of Psychology
        55:591--621,
        \url{https://doi.org/10.1146/annurev.psych.55.090902.142015}
        
    \bibitem{Belanche20}
        Belanche, D, Casal\'{o}, LV, Flavi\'{a}n, C, Schepers, J
        (2020)
        Service robot implementation: a theoretical framework and research agenda.
        The Service Industries Journal
        40(3--4): 203--225,
        \url{https://doi.org/10.1080/02642069.2019.1672666}
        
    \bibitem{Xiao19}
        Xiao, L, Kumar, V
        (2019)
        Robotics for Customer Service: A Useful Complement or an Ultimate Substitute?.
        Journal of Service Research
        24(1): 9--29,
        \url{https://doi.org/10.1177/1094670519878881}
        
    \bibitem{Ham15}
        Ham, J, Cuijpers, RH, Cabibihan, JJ
        (2015)
        Combining Robotic Persuasive Strategies: The Persuasive Power of a Storytelling Robot that Uses Gazing and Gestures.
        International Journal of Social Robotics
        7: 479--487,
        \url{https://doi.org/10.1007/s12369-015-0280-4}
        
    \bibitem{Chidambaram12}
        Chidambaram, V, Chiang, Y, Mutlu, B
        (2012)
        Designing Persuasive Robots: How Robots Might Persuade People Using Vocal and Nonverbal Cues.
        In: ACM/IEEE International Conference on Human-Robot Interaction
        293--300,
        \url{https://doi.org/10.1145/2157689.2157798}
        
    \bibitem{Cialdini01}
        Cialdini, RB
        (2001)
        Harnessing the Science of Persuasion.
        Harvard Business Review
        79(9): 72--79
        
    \bibitem{Andrist13}
        Andrist, S, Spannan, E, Mutlu, B
        (2013)
        Rhetorical robots: Making robots more effective speakers using linguistic cues of expertise.
        In: ACM/IEEE International Conference on Human-Robot Interaction
        341--348,
        \url{https://doi.org/10.1109/HRI.2013.6483608}
        
    \bibitem{Saunderson22}
        Saunderson, S, Nejat, G
        (2022)
        Investigating Strategies for Robot Persuasion in Social Human–Robot Interaction.
        IEEE Transactions on Cybernetics
        52(1): 641--653,
        \url{https://doi.org/10.1109/TCYB.2020.2987463}
        
    \bibitem{Bertacchini17}
        Bertacchini, F, Bilotta, E, Pantano, P
        (2017)
        Shopping with a robotic companion.
        Computers in Human Behavior
        77: 382--395,
        \url{https://doi.org/10.1016/j.chb.2017.02.064}
        
    \bibitem{Asch56}
        Asch, SE
        (1956)
        Studies of independence and conformity: I. A minority of one against a unanimous majority.
        Psychological Monographs: General and Applied
        70(9): 1–-70,
        \url{https://doi.org/10.1037/h0093718}
        
    \bibitem{Salomons17}
        Salomons, N, Sebo, SS, Qin, M, Scassellati, B
        (2017)
        A Minority of One against a Majority of Robots: Robots Cause Normative and Informational Conformity.
        ACM Transactions on Human-Robot Interaction
        10(2): 15,
        \url{https://doi.org/10.1145/3442627}
        
    \bibitem{Heider58}
        Heider, F
        (1958)
        The psychology of interpersonal relations.
        New York
        John Wiley \& Sons.
        \url{https://doi.org/10.1037/10628-000}
        
    \bibitem{Kadowaki08}
        Kadowaki, K, Kobayashi, K, Kitamura, Y
        (2008)
        Influence of Social Relationships on Multiagent Persuasion.
        In: 7th international joint conference on Autonomous agents and multiagent systems
        1221--1224
        
    \bibitem{Tae21}
        Tae, MI, Ogawa, K, Yoshikawa, Y, Ishiguro, H
        (2021)
        Using Multiple Robots to Increase Suggestion Persuasiveness in Public Space.
        Applied Science
        11(13): 6080,
        \url{https://doi.org/10.3390/app11136080}

    \bibitem{Okeefe90}
        O'Keefe, DJ
        (1990)
        Persuasion: Theory and Research.
        Newbury Park
        CA: Sage Publications.
        \url{https://doi.org/10.1177/002194369202900107}

    \bibitem{Haiman49}
        Haiman, FS
        (1949)
        An experimental study of the effects of ethos in public speaking.
        Speech Monographs
        16(2): 190-202,
        \url{https://doi.org/10.1080/03637754909374974}

    \bibitem{Peng19}
        Peng, Z, Kwon, Y, Lu, J, Wu, Z, Ma, X
        (2019)
        Design and Evaluation of Service Robot's Proactivity in Decision-Making Support Process.
        In: CHI Conference on Human Factors in Computing Systems
        98: 1–13,
        \url{https://doi.org/10.1145/3290605.3300328}
        
    \bibitem{Carolis17}
        Carolis, BD, Gemmis, M, Lops, P, Palestra, G
        (2017)
        Recognizing users feedback from non-verbal communicative acts in conversational recommender systems.
        Pattern Recognition Letters
        99: 87--95,
        \url{https://doi.org/10.1016/j.patrec.2017.06.011}
        
    \bibitem{Woiceshyn17}
        Woiceshyn, L, Wang, Y, Nejat, G, Benhabib, B
        (2017)
        Personalized clothing recommendation by a social robot.
        In: International Symposium on Robotics and Intelligent Sensors
        179--185,
        \url{https://doi.org/10.1109/IRIS.2017.8250118}
        
    \bibitem{Alslaity19}
        Alslaity, A, Tran, T
        (2019)
        Towards Persuasive Recommender Systems.
        In: IEEE International Conference on Information and Computer Technologies
        143--148,
        \url{https://doi.org/10.1109/INFOCT.2019.8711416}
        
    \bibitem{Okafuji21}
        Okafuji, Y, et al
        (2021)
        Persuasion Strategies for Social Robot to Keep Humans Accepting Daily Different Recommendations.
        In: IEEE/RSJ International Conference on Intelligent Robots and Systems
        1950--1957,
        \url{https://doi.org/10.1109/IROS51168.2021.9636772}
        
    \bibitem{Shiomi13}
        Shiomi, M, et al
        (2013)
        Recommendation Effects of a Social Robot for Advertisement-Use Context in a Shopping Mall.
        International Journal of Social Robotics
        143--148,
        \url{https://doi.org/10.1109/INFOCT.2019.8711416}
        
    \bibitem{Okafuji22}
        Okafuji, Y, at al
        (2022)
        Behavioral assessment of a humanoid robot when attracting pedestrians in a mall.
        International Journal of Social Robotics
        14: 1731--1747
        \url{https://doi.org/10.1007/s12369-022-00907-9}
        
    \bibitem{Tonkin17}
        Tonkin, M,et al
        (2017)
        Would you like to sample? Robot engagement in a shopping centre.
        In: IEEE International Symposium on Robot and Human Interactive Communication
        42--49,
        \url{https://doi.org/10.1109/ROMAN.2017.8172278}
        
    \bibitem{Herse18b}
        Herse, S, et al
        (2018)
        Bon Appetit! Robot Persuasion for Food Recommendation.
        In: ACM/IEEE International Conference on Human-Robot Interaction
        125–-126,
        \url{https://doi.org/10.1145/3173386.3177028}
        
    \bibitem{Sakai22}
        Sakai, K, Nakamura, Y, Yoshikawa, Y, Ishiguro, H
        (2022)
        Effect of Robot Embodiment on Satisfaction With Recommendations in Shopping Malls.
        IEEE Robotics and Automation Letters,
        7(1): 366--372,
        \url{https://doi.org/10.1109/LRA.2021.3128233}
        
    \bibitem{Niemela17a}
        Niemel\"{a}, M, Heikkil\"{a}, P, Lammi, H
        (2017)
        A Social Service Robot in a Shopping Mall: Expectations of the Management, Retailers and Consumers.
        In: ACM/IEEE International Conference on Human-Robot Interaction
        227--228,
        \url{https://doi.org/10.1145/3029798.3038301}

    \bibitem{Riek12}
        Riek, LD
        (2012)
        Wizard of Oz studies in HRI: a systematic review and new reporting guidelines.
        Journal of Human-Robot Interaction
        1(1): 119--136,
        \url{https://doi.org/10.5898/JHRI.1.1.Riek}
        
    \bibitem{Fay11a}
        Fay, MJ
        (2011)
        Informal communication of co-workers: a thematic analysis of messages.
        Qualitative Research in Organizations and Management
        6(3): 212--229,
        \url{https://doi.org/10.1108/17465641111188394}
        
    \bibitem{Fay11b}
        Fay, MJ, Lkine, SL
        (2011)
        Coworker Relationships and Informal Communication in High-Intensity Telecommuting.
        Journal of Applied Communication Research 
        39(2): 144--163,
        \url{https://doi.org/10.1080/00909882.2011.556136}
        
    \bibitem{Alparslan15}
        Alparslan, AM, Kilinc, U
        (2015)
        The power of informal communication and perceived organizational support on energy at work and extra-role behavior: A survey on teachers.
        Journal of Human Sciences
        12(2): 113--138,
        \url{https://doi.org/10.14687/ijhs.v12i2.3243}
        
    \bibitem{Heerink10}
        Heerink, M, Kr\"{o}se, B, Evers, V, Wielinga, B 
        (2010)
        Assessing Acceptance of Assistive Social Agent Technology by Older Adults: the Almere Model.
        International Journal of Social Robotics
        2: 361--375,
        \url{https://doi.org/10.1007/s12369-010-0068-5}

    \bibitem{Chan19}
        Chan, APH, Tung, VWST
        (2019)
        Examining the effects of robotic service on brand experience: the moderating role of hotel segment.
        Journal of Travel \& Tourism Marketing 
        36(4): 458--468,
        \url{https://doi.org/10.1080/10548408.2019.1568953}
        
    \bibitem{Bennett17}
        Bennett, M, Williams, T, Thames, D, Scheutz, M
        (2017)
        Differences in interaction patterns and perception for teleoperated and autonomous humanoid robots.
        In: IEEE/RSJ International Conference on Intelligent Robots and Systems
        6589--6594,
        \url{https://doi.org/10.1109/IROS.2017.8206571}
        
    \bibitem{Kuwamura12}
        Kuwamura, K, Minato, T, Nishio, S, Ishiguro, H
        (2012)
        Personality distortion in communication through teleoperated robots.
        In: IEEE International Symposium on Robot and Human Interactive Communication
        49--54,
        \url{https://doi.org/10.1109/ROMAN.2012.6343730}
        
    \bibitem{Hart88}
        Hart, SG, Staveland, LE
        (1988)
        Development of NASA-TLX (Task Load Index): Results of Empirical and Theoretical Research.
        Advances in Psychology
        52: 139--183,
        \url{https://doi.org/10.1016/S0166-4115(08)62386-9}
        
    \bibitem{Zajonc68}
        Zajonc, RB
        (1968)
        Attitudinal effects of mere exposure.
        Journal of Personality and Social Psychology
        9(2): 1--27,
        \url{https://doi.org/10.1037/h0025848}
        
    \bibitem{Niemela17b}
        Niemel\"{a}, M, Arvola, A, Aaltonen, I
        (2017)
        Monitoring the Acceptance of a Social Service Robot in a Shopping Mall: First Results.
        In: ACM/IEEE International Conference on Human-Robot Interaction
        225--226,
        \url{https://doi.org/10.1145/3029798.3038333}
        
    \bibitem{SongCS22}
        Song, CS, Kim, Y-K
        (2022)
        The role of the human-robot interaction in consumers' acceptance of humanoid retail service robots.
        Journal of Business Research
        146: 489--503,
        \url{https://doi.org/10.1016/j.jbusres.2022.03.087}
        
    \bibitem{Lu19}
        Lu, L, Cai, R, Gursoy, D
        (2019)
        Developing and validating a service robot integration willingness scale.
        International Journal of Hospitality Management
        80: 36--51,
        \url{https://doi.org/10.1016/j.ijhm.2019.01.005}
        
    \bibitem{Qiu20}
        Qiu, H, Li, M, Shu, B, Bai, B
        (2020)
        Enhancing hospitality experience with service robots: the mediating role of rapport building.
        Journal of Hospitality Marketing \& Management
        29(3): 247--268,
        \url{https://doi.org/10.1080/19368623.2019.1645073}
        
    \bibitem{Ajoudani18}
        Ajoudani, A, et al
        (2018)
        Progress and prospects of the human–robot collaboration.
        Autonomous Robotics
        42: 957--975,
        \url{https://doi.org/10.1007/s10514-017-9677-2}
        
    \bibitem{Welfare19}
        Welfare, KS, Hallowell, MR, Shah, JA, Riek, LD
        (2019)
        Consider the Human Work Experience When Integrating Robotics in the Workplace.
        In: ACM/IEEE International Conference on Human-Robot Interaction
        75--84,
        \url{https://doi.org/10.1109/HRI.2019.8673139}
        
    \bibitem{Ferreira14}
        Ferreira, P, Doltsinis, S, Lohse, N
        (2014)
        Symbiotic Assembly Systems –- A New Paradigm.
        Procedia CIRP
        17: 26--31,
        \url{https://doi.org/10.1016/j.procir.2014.01.066}
        
    \bibitem{McClure18}
        McClure PK
        (2018)
        ``You're Fired,'' Says the Robot: The Rise of Automation in the Workplace, Technophobes, and Fears of Unemployment.
        Social Science Computer Review
        36(2): 139--156,
        \url{https://doi.org/10.1177/0894439317698637}
        
    \bibitem{McCartney20}
        McCartney, G, McCartney, A
        (2020)
        Rise of the machines: towards a conceptual service-robot research framework for the hospitality and tourism industry.
        International Journal of Contemporary Hospitality Management
        32(12): 3835--3851,
        \url{https://doi.org/10.1108/IJCHM-05-2020-0450}
        
    \bibitem{Niemela19}
        Niemel\"{a}, M, Heikkil\"{a}, P, Lammi, H, Oksman, V
        (2019)
        A Social Robot in a Shopping Mall: Studies on Acceptance and Stakeholder Expectations.
        Social Robots: Technological, Societal and Ethical Aspects of Human-Robot Interaction
        Switzerland
        Springer Nature.
        119--144,
        \url{ https://doi.org/10.1007/978-3-030-17107-0\_7}
        
    \bibitem{Niemela17c}
        Niemel\"{a}, M, Heikkil\"{a}, P, Lammi, H, Oksman, V
        (2017)
        Shopping Mall Robots –- Opportunities and Constraints from the Retailer and Manager Perspective.
        In: International Conference on Social Robotics
        485--494,
        \url{https://doi.org/10.1007/978-3-319-70022-9\_48}
        
    \bibitem{Shi16}
        Shi, C, Satake, S, Kanda, T, Ishiguro, H
        (2016)
        How would store managers employ social robots?.
        In: ACM/IEEE International Conference on Human-Robot Interaction
        519--520,
        \url{https://doi.org/10.1109/HRI.2016.7451835}
        
    \bibitem{Graaf16}
        de Graaf MA, Allouch SB, van Dijk JAGM
        (2016)
        Long-term evaluation of a social robot in real homes.
        Interaction Studies
        17(3):1--25,
        \url{https://doi.org/10.1075/is.17.3.08deg}

  \end{thebibliography}


\end{document}